\documentclass[10pt, a4paper]{article}

\usepackage[]{lrec-coling2024} 
\usepackage{multicol}
\usepackage{graphicx}
\usepackage{microtype}
\usepackage{amsfonts}
\usepackage{mathrsfs}
\usepackage{multirow}
\usepackage{amssymb}
\usepackage{bbding}
\usepackage{subfigure} 
\usepackage{CJKutf8}
\usepackage{colortbl}
\usepackage{color}
\usepackage{booktabs}
\usepackage{makecell}
\usepackage{calligra}
\usepackage{algorithm}
\usepackage{algorithmicx}
\usepackage{algpseudocode} 
\usepackage{amsmath} 

\title{CHisIEC: An Information Extraction Corpus for Ancient Chinese History}

\name{Xuemei Tang$^{1,2}$, Zekun Deng$^{1,2}$,  Qi Su$^{3,2,4}$, Hao Yang$^{2,4}$, Jun Wang$^{1,2,4}$ \sthanks{Corresponding author}} 

\address{$^1$ Department of Information Management, Peking University \\$2$ Research Center for Digital Humanities of Peking University \\ $^3$ School of Foreign Languages, Peking University\\ $^4$ Institute for Artificial Intelligence, Peking University\\
\texttt{tangxuemei@stu.pku.edu.cn}\\ \texttt{\{dzk, sukia, junwang, yanghao2008\}@pku.edu.cn}}

\abstract{ 
Natural Language Processing (NLP) plays a pivotal role in the realm of Digital Humanities (DH) and serves as the cornerstone for advancing the structural analysis of historical and cultural heritage texts. This is particularly true for the domains of named entity recognition (NER) and relation extraction (RE). In our commitment to expediting ancient history and culture, we present the ``Chinese Historical Information Extraction Corpus''(CHisIEC). CHisIEC is a meticulously curated dataset designed to develop and evaluate NER and RE tasks, offering a resource to facilitate research in the field. Spanning a remarkable historical timeline encompassing data from 13 dynasties spanning over 1830 years, CHisIEC epitomizes the extensive temporal range and text heterogeneity inherent in Chinese historical documents. The dataset encompasses four distinct entity types and twelve relation types, resulting in a meticulously labeled dataset comprising 14,194 entities and 8,609 relations. To establish the robustness and versatility of our dataset, we have undertaken comprehensive experimentation involving models of various sizes and paradigms. Additionally, we have evaluated the capabilities of Large Language Models (LLMs) in the context of tasks related to ancient Chinese history. The dataset and code are available at \url{https://github.com/tangxuemei1995/CHisIEC}. \\
\newline \Keywords{Ancient Chinese History, Dataset Annotation, Named Entity Recognition, Relation Extraction}}

\begin{document}
\begin{CJK*}{UTF8}{gbsn}
\maketitleabstract

\section{Introduction}


Historical and cultural heritage preservation is an important branch of digital humanities, where the rich tapestry of the past meets the cutting-edge tools of the digital age. This field has been significantly enhanced by applying various technologies, including Natural Language Processing (NLP), Computer Vision (CV), and Knowledge Graphs (KG). 

In recent works, many studies have endeavored to structure cultural heritage and historical documents. For example, ~\citet{Kim_Kim_Son_Lim_2022} annotated a mixed multilingual corpus of Korean cultural heritage related to entities.
In addition, some historical documents, such as newspapers and periodicals, have also received attention. ~\citet{neudecker-2016-open} and ~\citet{Ehrmann_Romanello_Fluckiger_Clematide} focused on the entity annotation and recognition of European historical newspapers. ~\cite{Bekele_De_By_Singh_2016} extracted the spatial and temporal entities from the Brazilian historic expedition gazetteer. Moreover, \citet{Nundloll_Smail_Stevens_Blair_2022} identified custom entities and domain entities from the annals of a historical Botany journal.

The cornerstone of advancing the automatic Information Extraction (IE) models in this field lies in the availability of labeled data. However, the domain of ancient Chinese historical documents presents a unique challenge due to the extensive time span they encompass and the linguistic heterogeneity they exhibit. Although ~\citet{Li_Wei_Liu_Chen_Fang_Jiang_2021} and ~\citet{jizijing2021} attempted to build a corpus of information extraction based on ancient Chinese historical documents, there are only 1,600 and 4,000 pieces of data, respectively. Unfortunately, these corpora are still significantly undersized to serve as a solid foundation for developing deep learning models. This limitation greatly hinders the implementation of IE techniques in the ancient Chinese historical documents domain.


To tackle the challenges associated with information extraction from ancient Chinese historical documents, we introduce an ancient Chinese Historical Information Extraction Corpus (CHisIEC), which is a high-quality specialized dataset for ancient Chinese historical documents and can be used for NER and RE tasks. In constructing this dataset, considering the large time span of ancient Chinese history, we select 13 historical books from the representative \textit{Twenty-Four Histories} as the raw data, spanning over 1830 years. Then, we further combine the content and linguistic characteristics of the historical documents, define specific entity types and relation types, and craft detailed annotation guidelines. Finally, we invite annotators to annotate according to these guidelines to create the annotated dataset.

The contributions of this paper are as follows.
\begin{itemize}
\item We construct an information extraction dataset for ancient Chinese historical documents, which is the largest available and contains both NER and RE tasks. Our dataset includes more than 130K tokens, 14,194 entities, and 8,609 relations.

\item The data in the corpus come from 13 historical books spanning up to 1830 years, preserving the characteristics of ancient Chinese historical documents in terms of time span and text heterogeneity.

\item We validate the utility of our dataset by conducting comprehensive experiments using models of varying sizes and within different paradigms, including pre-trained language models (PLMs) and large language models (LLMs).
\end{itemize}

\section{Related Work}

\subsection{Historical Dataset}

Texts in history and cultural heritage exhibit heterogeneity and noise due to their association with diverse time periods, domains, and the influence of OCR results. Several works have attempted to preserve these data features in annotated data. 
For instance, ~\citet{Kim_Kim_Son_Lim_2022} introduced an entity-related Korean cultural heritage corpus KoCHET, which encompasses three sub-tasks: NER, RE, and entity typing (ET). The raw data for this corpus is sourced from e-museum digitized data in both Korean and Chinese; \citet{Nundloll_Smail_Stevens_Blair_2022} annotated custom entities such as people, nationalities, buildings, organizations, countries, times, and events in the Journal of Historical Botany, as well as domain entities such as plant names, observers, locations, spatial relationships, topographic attributes, and abundance.

Newspapers serve as typical historical sources and form the foundational material for creating historical datasets. For instance, 
~\citet{Ehrmann_Romanello_Fluckiger_Clematide} released the entity dataset HIPE, designed for evaluating named entity processing in French, German, and English historical newspapers. The dataset includes tasks related to entity recognition, classification, and linking, with corpus texts originating from newspapers spanning from 1798 to 2018. Additionally, ~\citet{neudecker-2016-open} produced a corpus of 400 Dutch/French/German newspaper pages that were manually filtered and annotated with named entities such as people, locations, and organizations. The corpus consists mainly of pre-1900 newspaper texts with historical spelling variations.

In the field of Chinese history and cultural heritage, there are publicly available datasets. ~\citet{Zinin_Xu} created a historical lexicon and semantic corpus named CCDH, utilizing open-source \textit{Twenty-four Histories}, which consists of Classical Chinese gender-specific terms. CCDH supports both synchronic and asynchronous studies of gender terms in ancient Chinese.
In the domains of NER and RE, ~\citetlanguageresource{Li_Wei_Liu_Chen_Fang_Jiang_2021} constructed a few-shot ancient Chinese relation dataset (TinyACD-RC) containing 1,600 instances and 32 relation types. Additionally, ~\citet{jizijing2021} developed a RE corpus with 25 relation types and 4413 samples based on \textit{Twenty-Four Histories}.

\subsection{Information Extraction}

Information extraction is the foundation of NLP systems and aims to extract structured information from unstructured or semi-structured data sources automatically. In recent years, deep learning methods have made achievements in information extraction tasks~\cite{Wu_He_2019, Nguyen_Min_Dernoncourt_Nguyen_2022, Liang_Wu_Li_Li_2022}, these methods are categorized into two types, one is to divide the information extraction into multiple sub-tasks, and model the multiple sub-tasks separately, such as named entity recognition, relation classification, event triggering detection, and event argument classification. Another category is modeling IE as a joint task, e.g. named entity recognition and relation classification are often modeled as a joint task. For the joint task of RE, different modeling paradigms have been proposed, such as machine reading comprehension-based approach~\cite{Li_Yin_Sun_Li_Yuan_Chai_Zhou_Li_2019, Zhao_Yan_Cao_Li_2020}, sequence labeling-based approach~\cite{Zheng_Wang_Bao_Hao_Zhou_Xu_2017}, span-based approach~\cite{Eberts_Ulges_2021, Ji_Yu_Li_Ma_Wu_Tan_Liu_2020}, and generation-based approach~\cite{Huguet_Cabot_Navigli_2021, Nayak_Ng_2020}.


Recently, the development of large language models, such as GPT-3~\cite{Brown_2020}, ChatGPT~\cite{Ouyang_Wu_Jiang_Almeida_Wainwright_Mishkin_Zhang_Agarwal_Slama_Ray_et_2022}, and GPT-4~\footnote{https://openai.com/research/gpt-4}, has significantly advanced the field of natural language understanding and generation. These LLMs have been trained on massive text corpora to generate coherent and contextually accurate text.

Instruction tuning~\cite{Lou_Zhang_Yin_2023} is a novel paradigm for using natural language instructions to guide LLMs to complete downstream tasks, and it shows great promise for observing the generalization of task sets. Some works  ~\cite{Gui_Zhang_Ye_Zhang_2023, Wang_Zhou_Zu_Xia_Chen_Zhang_Zheng_Ye_Zhang_Gui_et_2023} tried to transfer the IE task samples to instruction-formatted instances, then fine-tune LLMs in a supervised learning way~\cite{Zhao_Zhou_Li_Tang_Wang_Hou_Min_Zhang_Zhang_Dong_et_2023}.

Recent studies on LLMs such as GPT-3~\cite{Brown_2020} have shown that LLMs perform well in a variety of downstream tasks without any training or fine-tuning, only formulating the task description and demonstrations in the form of natural language text as instructions, which is known as in-context learning ~\cite{Zhao_Zhou_Li_Tang_Wang_Hou_Min_Zhang_Zhang_Dong_et_2023}. Many studies have achieved information extraction by adapting the in-context learning strategies. For example, ~\citet{Wei_Cui_Cheng_Wang_Zhang_Huang_Xie_Xu_Chen_Zhang_et_2023} convert the IE task into a multi-turn question-answering task, and then get structured data by asking ChatGPT in two-stages; ~\cite{Ling_Zhao_Zhang_Liu_Cheng_Wang_Chen_Osaki_Matsuda_Chen_2023} added an error correction mechanism to enhance the confidence of the generated relations.

\begin{figure*}[t]
    \centering
    \subfigure[Entity types]{ \includegraphics[width=7cm,height=5cm]{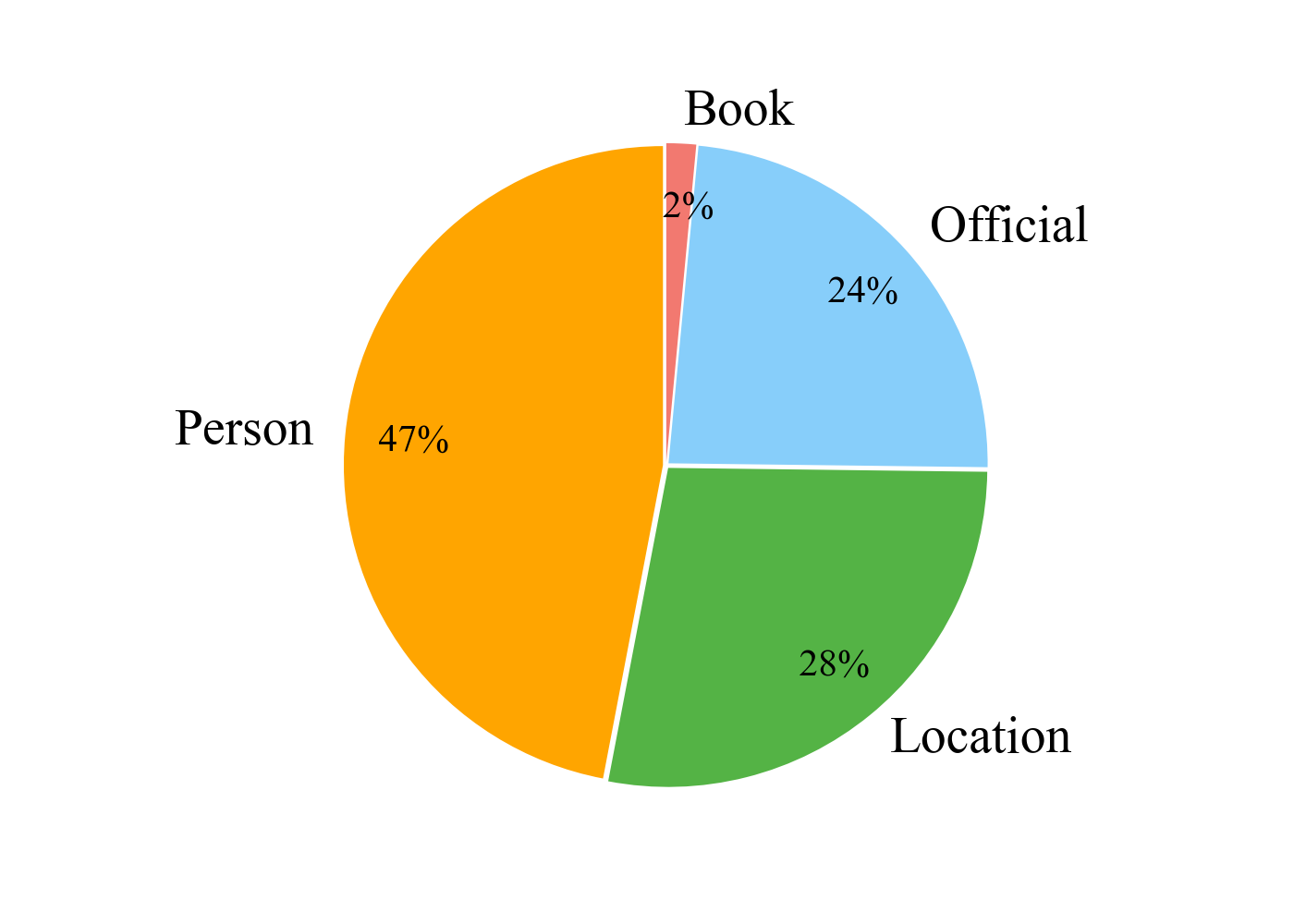}
    \label{ner_pie}
    }
    \quad
    \subfigure[Relation types]{ \includegraphics[width=8cm,height=5cm]{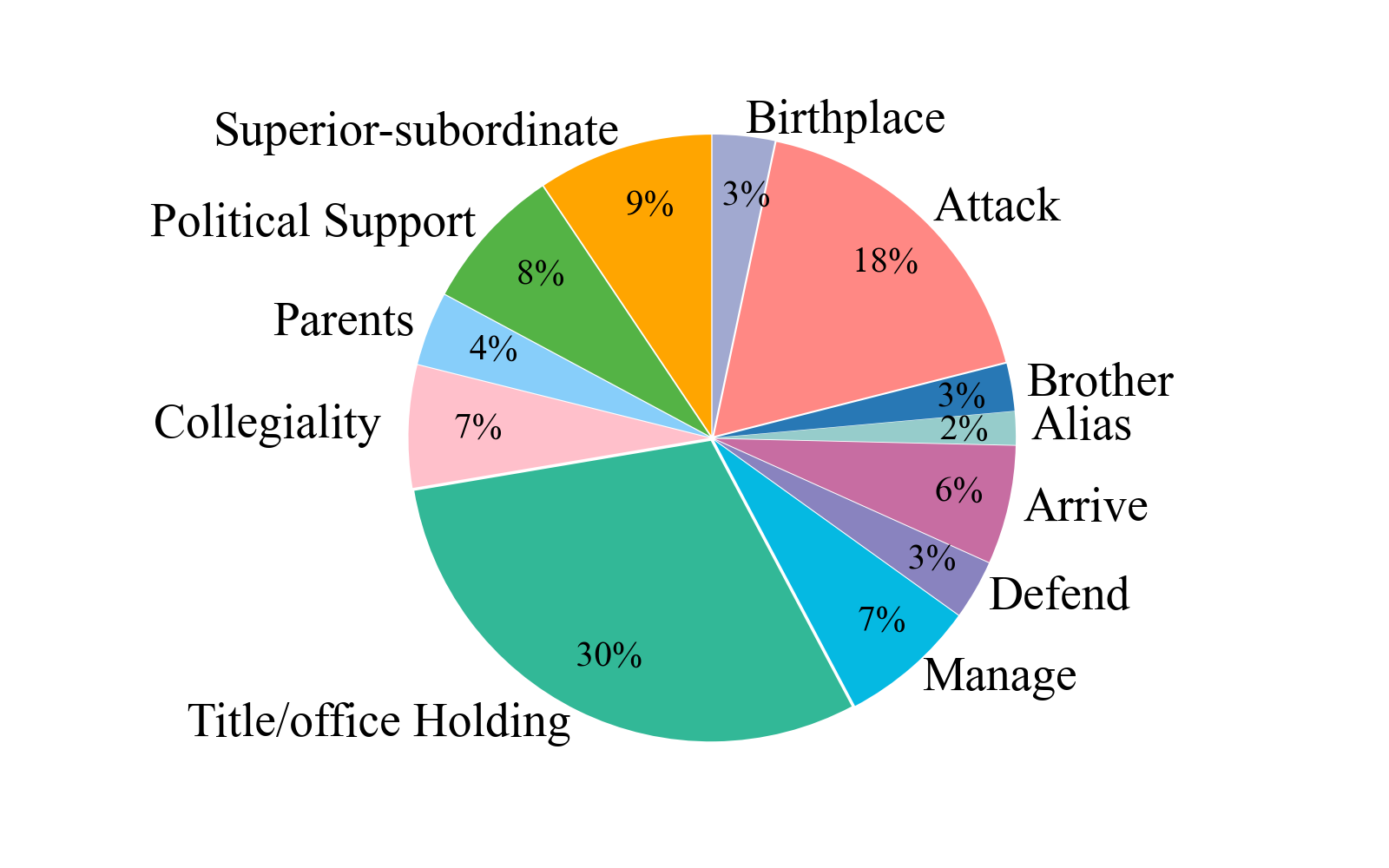}
    \label{re_pie}}
    \caption{Visualization of labels as a percentage.}
    \label{nerlabel}
\end{figure*}

\section{Corpus Annotation}

 
CHisIEC is a specialized corpus designed for the study of ancient Chinese history. The raw data for this corpus is sourced from the \textit{Twenty-four Histories}, a compilation of the official histories of the Chinese twenty-four dynasties. The \textit{Twenty-four Histories}, also known as the ``Official History'' chronicle over 4,000 years of Chinese history. It consists of 3,213 volumes containing approximately 40 million tokens.
 

We select 22 volumes from 13 books within the \textit{Twenty-four Histories} as the texts to be labeled, including \textit{The Records of the Grand Historian}, \textit{The Book of Han}, \textit{The Book of Tang}, \textit{The History of Song}, \textit{The History of Ming}, and so on. These texts span over 1830 years of Chinese history. Subsequently, we randomly divided the text into segments, each approximately 100 tokens in length, resulting in a total of 150K characters.

\subsection{Analysis on CHisIEC}

\subsubsection{Statistics}
 
\begin{table}[h]
    \centering
    \setlength{\tabcolsep}{0.5mm}
    \begin{tabular}{c|c|c|c}
    \hline
      \hline
    \textbf{Dataset} &   \textbf{Samples} &  \textbf{Characters} &\textbf{Entities/Relations}   \\
        \hline
         NER  &   2,185  & 135,713  &14,194 \\
         \hline
         RE  &   1,948 & 88,177  & 8609\\

         \hline
           \hline
    \end{tabular}
    \caption{Statistics of CHisIEC dataset.}
    \label{corpus}

\end{table}
As mentioned earlier, our raw data consists of 150K characters. After annotation, the statistics of the corpus data are presented in Table~\ref{corpus}. We filter out samples with no entities in the raw data, resulting in 2,185 samples with 135K characters, and we labeled a total of 14,194 entities, forming the named entity dataset. Using the NER dataset as a foundation, we further screen samples without relation annotations to create the RE dataset, comprising 1,948 samples with 88K characters and a total of 8,609 triplets. To facilitate model training, we divided both datasets into training, validation, and test sets in an 8:1:1 ratio. 

We plot the category occupancy of the two datasets as shown in Figure~\ref{nerlabel}.
As we can see in Figure~\ref{ner_pie}, the entity types in the dataset exhibit an uneven distribution, with a significant number of person, place, and official entities, while book entities are relatively scarce, accounting for only 2\% of the dataset. This suggests that persons are a prominent focus in ancient Chinese historical documents, whereas books play a less central role in the narratives found in the \textit{Twenty-four Histories}.

In Figure~\ref{re_pie}, we observe that the most frequent relation type is the Title/Office Holding relation, followed by the Attack relation, with the Superior-subordinate relation also demonstrating a relatively high frequency. All other relation types have frequencies below 800. The relation types that accounted for the lowest percentage were the Alias relation, the Birthplace relation, the Defend relation, and the Brother relation. These frequency patterns align with the predominant themes of political and military topics in ancient Chinese historical texts.

\subsubsection{Linguistic Analysis}

\begin{figure*}[t]
    \centering
    \includegraphics[width=15cm,height=5cm]{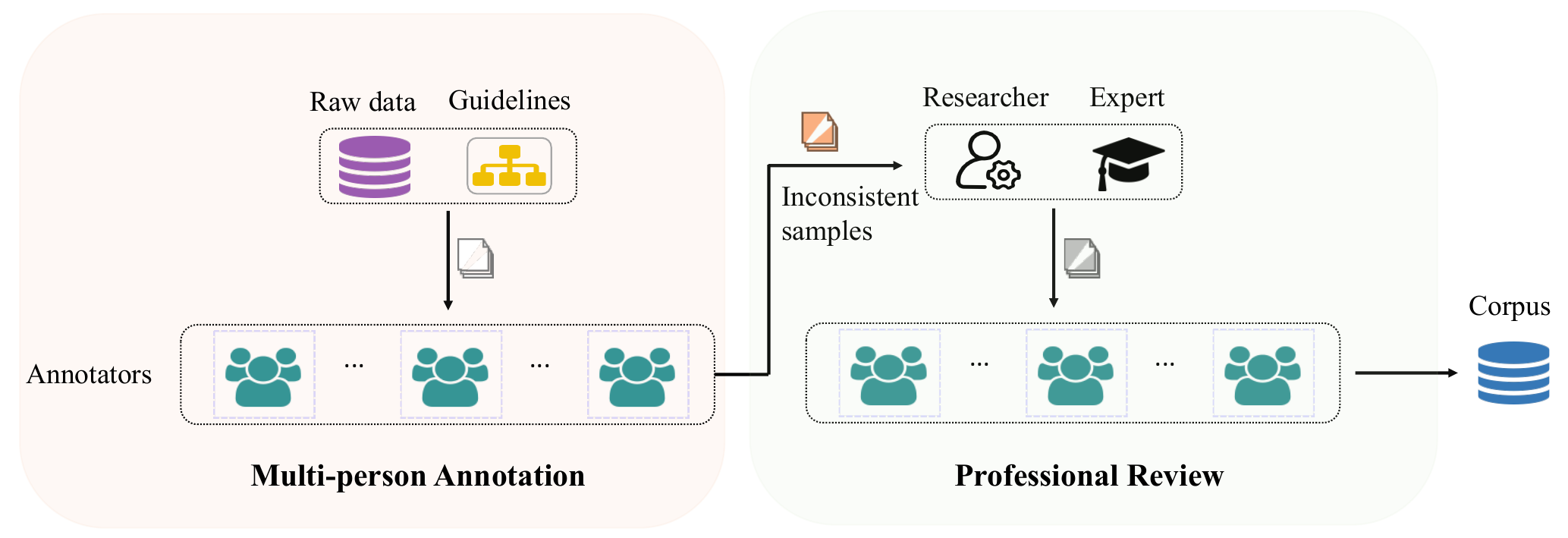}
    \caption{Annotation process for our corpus.}
    \label{annotation_process}
\end{figure*}

In this section, we analyze the linguistic features of the annotated corpus.

First, the corpus has a long time span. It is based on the official histories of 13 dynasties, with the earliest text from \textit{The Records of the Grand Historian} dating back to about 91 B.C. and the latest from \textit{The History of Ming} in 1739 A.D. This resulted in a remarkable period of 1830 years.

Second, the long time span leads to high heterogeneity within the corpus. The language used in the corpus is ancient Chinese, which differs significantly from modern Chinese in vocabulary and grammar. Ancient Chinese is categorized into three developmental stages: Early Ancient, Middle Ancient, and Near Ancient, each with distinct linguistic features. Our dataset primarily contains texts from the latter two periods, contributing to the corpus's high heterogeneity.

Third, the corpus exhibits exceptionally high linguistic information density. While modern Chinese is known for its information-dense nature, ancient Chinese surpasses it in this aspect. This heightened information density is evident in the annotated data. Based on the statistics presented in Table~\ref{corpus}, on average, each sample contains six entities and four pairs of triplets.

\subsection{Annotation Process}

In this section, we will provide an overview of the annotation process and the annotation schema.

\textbf{Annotation mode}. In the practice of large-scale data annotation, we adopt the mode of ``multi-person annotation'' and ``professional review'', the annotation process is shown in Figure~\ref{annotation_process}. Initially, we recruit 18 undergraduate students as annotators, dividing them into six groups, with an equal distribution of data to each group. Within these groups, annotators independently annotate the same text based on the provided annotation guidelines.
Subsequently, any inconsistencies in the annotations within each group are identified, and resolutions are determined through discussions involving both a task researcher and an expert with a historical background. After these discussions, the annotators perform a second round of proofreading on the initial annotations, incorporating the suggestions provided by the experts.

\textbf{Annotation guidelines}. We design the following annotation rules to ensure the annotation consistency among the annotators.

\begin{itemize}
    \item Entity type annotation is context-dependent. Certain words can function as both personal names and official positions. For example, ``繇王不能矫其众持正。 (Yao King is not able to correct his people.)'', where ``繇王 (Yao King)'' is an official position, but due to the context in the sentence, it is labeled as a person.
    
    \item When labeling entities, fine-grained spans take precedence over coarse-grained spans. For example, in the sentence ``闽越王无诸及越东海王摇者。 (King of Minyue, Wuzhu, and the King of Yuedonghai, Yao)'', ``闽越王 (King of Minyue)'' and ``无诸 (Wuzhu)'' are co-referential and are annotated as separate entities. 

    \item Relation annotation is determined by context. In certain cases, there is a semantic overlap between relations, such as Collegiality and Superior-subordinate. Consequently, if the context unambiguously suggests a Superior-subordinate relationship, it will be labeled as such; if only two people are mentioned as working together, the relationship is labeled as Collegiality.
\end{itemize}

\subsection{Schema for Task Annotation}

Before corpus annotation, we formulate annotation specifications for different types of named entities and relations in ancient Chinese historical texts. We constantly update and improve the specifications in the annotation practice to make them more suitable for annotating ancient Chinese historical texts of different periods.

\subsubsection{Named Entity Type}

Establishing a named entity taxonomy is essential for annotating Chinese historical texts. Drawing from Chinese historiography, there are four pivotal facets for comprehending history: ``Catalog Studies'', ``Chronology Studies'', ``Historical Geography'', and ``Official Systems''. Catalog studies focus on book-related matters, chronology studies delve into the timing of events and materials, historical geography explores the locations of people's activities and events, and official systems research addresses changes within the administrative framework.

In line with this framework, we identify four primary entity categories: Person, Location, Official, and Book.

\textbf{Person}. As creators and witnesses of history, people are central in historical records across eras and dynasties, including emperors, politicians, cultural figures, generals, and other influential societal members. 

\textbf{Location}. Spatial context situates people and events, providing the stage and background for historical activities, decisions, and changes. Locations range from regions, capitals, counties, geographical features, and palaces.

\textbf{Official position}. Political figures commonly hold formal posts, with the evolution of official systems intertwined with the rise and fall of dynasties. Tracking offices illustrate sociopolitical characteristics of different periods.

\textbf{Book}. As vessels of thought and culture, books offer insights into academic advancements and biographical details of historical figures.

\subsubsection{Relation Type}

\textit{Twenty-Four Histories} are mainly concerned with politics, military, and culture. By combining expert knowledge with textual content analysis, we focus on five prominent types of relations: war, political, geopolitical, family, and personal attributes. 
As shown in Table~\ref{domain}, within each of these domains, we identify common relation types to capture connections between entities. We describe in detail the definition of each relation type as follows.
\begin{table}
    \begin{tabular}{c|c}

    \hline
        \textbf{Domain} &  \textbf{Relation Type}\\
        \hline
         \multirow{4}{*}{Politics} & Political Support,\\
         &Title/office Holding, \\
         &Collegiality, \\
         &Superior-subordinate\\
           \hline
         War & Attack, Defend\\
           \hline
         Geography & Arrive, Manage \\
           \hline
         Family & Parents, Brother\\
           \hline
         Personal Information &  Alias, Birthplace\\
           \hline
    \end{tabular}
    \caption{The history domains and the relation types.}
    \label{domain}
\end{table}

\begin{figure*}[t]
    \centering
    \includegraphics[width=13cm,height=4cm]{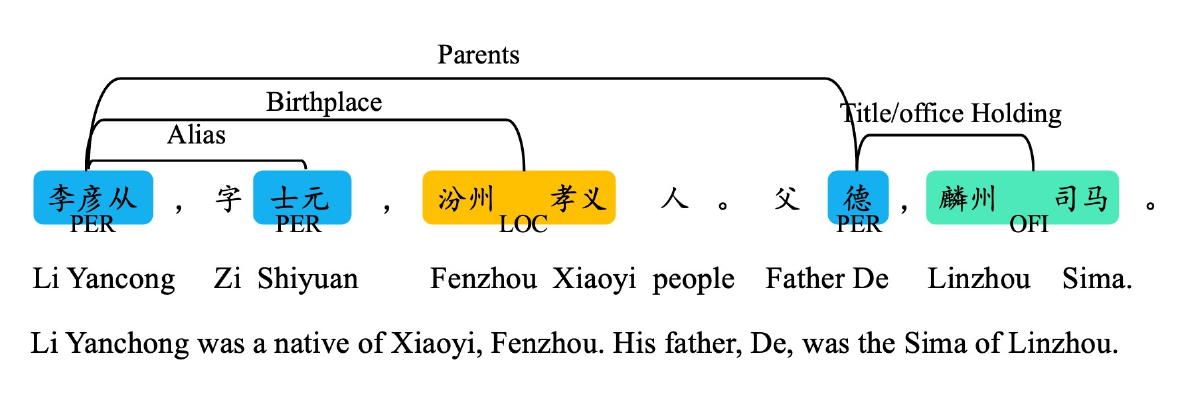}
    \caption{Illustration of an annotation sample.}
    \label{annotation_sample}
\end{figure*}

\begin{itemize}
    \item \textbf{Political Support}. 
    Political support is a significant and recurring theme in ancient Chinese historical texts. It can manifest in various forms, including support from subjects to rulers and interactions among political allies.
    This type of relation occurs between individuals, with the direction going from the supporter to the supported. 

    \item  \textbf{Title/office Holding}. Officials, posthumous titles, seals, and temple titles record the organizational structure of the ancient Chinese political and social system, the transmission of power, and the performance of official duties. The Title/Office Holding relation signifies that a person holds a specific position and title. This relation is directional, pointing from the person to the official position. 
    In ancient Chinese historical documents, the appearance of a character is usually accompanied by their position or title, making this type of relation particularly prevalent. 

    \item \textbf{Collegiality}. Collegiality describes the relationship between individuals who work in the same organization or institution, hold similar positions, or share similar status. In ancient Chinese historical documents, this relationship often signifies two or more people working for the same ruler or collaborating on a common task.  Collegiality is a non-directional relation. 

    \item \textbf{Superior-subordinate}. This relation pertains to two individuals who hold a superior-subordinate relationship. It involves texts that explicitly indicate someone as a superior or subordinate or contain actions implying such a relationship. The direction of this relation type is that the superior points to the subordinate.

    \item \textbf{Attack}. This type of relationship is closely linked to politics and warfare. Ancient Chinese historical documents contain a wealth of records about wars, encompassing conflicts between nations, tribes, political factions, and individuals. In our corpus, this relation primarily involves two individuals or a person and a location, with the direction going from the aggressor to the target. 

    \item \textbf{Defend}. This war-related relation is highly prevalent in ancient Chinese historical texts. It typically involves individuals leading armies stationed at specific locations for defensive purposes, a common military tactic in historical conflicts. This relation represents an action carried out by a person at a particular location. The direction of this relation goes from the person to the location. 

    \item \textbf{Manage}. This type of relation, categorized as a geopolitical relation, indicates that individuals are responsible for the management of specific locations. It often involves a person serving in a particular place and overseeing the state or county affairs. The direction of this relation goes from the person to the location. 

    \item \textbf{Arrive}. This is a geopolitical relation indicating that an individual arrives at a specific location. The direction of this relation goes from the person to the location. 

    \item \textbf{Birthplace}. In ancient Chinese historical texts, it was common for individuals to be introduced along with their place of birth. The direction of this relation goes from the person to the location. 

    \item \textbf{Parents}. In ancient China, there was a strong emphasis on blood ties, with the simplest and most direct blood relation being that of parents. The direction of this relation type goes from parents to children. 

    \item \textbf{Brother}. Brotherhood is one of the fundamental blood relations outside of parental relationships. In certain periods of China's social history, brother relations within the family could be intertwined with political relations. 
    In our corpus, the direction of the Brother relation points from the older brother to the younger brother

    \item \textbf{Alias}. In ancient China, people usually had aliases such as Zi (字) and Hao (号) in addition to their names. The direction of this relation goes from persons to their aliases.

\end{itemize}

We give an example of the annotation, as shown in Figure~\ref{annotation_sample}, where the annotator first annotates all the entities in the sentence and then annotates the relationships between the entities.

\begin{table*}[t]
\small
\centering
    \begin{tabular}{c|c|l}
    \hline
    \Xhline{1.pt}
    \multirow{11}{*}{NER} & \multirow{6}{*}{Instruction}& 你是一个实体识别工具，你需要识别出输入句子中的人物、地点、职官和书籍，\\
    &&输出格式为：人物：人物1，人物2；地点：地点1，地点2；职官：职官1，职官2；\\
    &&书籍：书籍1，书籍2。 (You are an entity recognition tool and you need to recognize \\
    &&persons, locations, officials, and books in an input sentence, and the output \\
    &&format: persons: person 1, person 2; locations: location 1, location 2; \\
   && officials: official 1, official 2; books: book 1, book 2.)\\
    \cline{2-3}
    &\multirow{3}{*}{Input}& 永泰元年，吐蕃请和，诏宰相元载、杜鸿渐与虏使者同盟。 (In the first year of the \\
    &&Yongtai era, Tubo asked for peace, and the Emperor ordered Chancellor Yuanzai and \\
    &&Du Hongjian to make an alliance with the captive's emissaries.)\\
   \cline{2-3}
    &\multirow{2}{*}{Output}& 人物：元载，杜鸿渐；地点：吐蕃；职官：宰相，使者；书籍：无。(Person: \\
    &&Yuan Zai, Du Hongjian; Locations: Tubo; Officials: chancellor, emissary; Books: none.)\\
    
    \hline
    \multirow{13}{*}{RE} & 
    \multirow{10}{*}{Instruction}& 你是一个语义抽取工具，现在已定义关系包括以下这些：敌对攻伐，任职，同僚，\\
    &&上下级，管理，驻守，到达，出生于某地，兄弟，父母，别名。以下句子中，\\
    && 实体由“【”， “】”标注出来，请你找出两个实体之间的关系。\\
    &&关系的方向由首实体指向尾实体，输出形式为：(首实体，关系，尾实体)。\\
    &&(You are a semantic extraction tool that now defined relations including the\\
    &&following Attack, title/office Holding, Collegiality, Superior-subordinate, Manage, Defend,  \\
    &&Arrive, Birthplace, Brother, Parents, Alias. In the following sentences, entities are marked\\
    &&by ``【'' and ``】''. You should identify the relation between the two entities. \\
    &&The direction of the relation is from the head entity to the tail entity and the output form is: \\
    &&(head entity, relation, tail entity).) \\
    \cline{2-3}
    & \multirow{2}{*}{Input}& 其年闰七月晦，李筠、【何福进】相率杀契丹帅【满达勒】。(On the 7th day of the leap \\
    &&month of this year, Li Yun and He Fujin killed the Khitan commander Mandalay.) \\
    \cline{2-3}
    &Output& (何福进，敌对攻伐，满达勒) (He Fujin, Attack, Mandalay)\\
    \hline
    \Xhline{1.pt}
    \end{tabular}
    \caption{Sample instructions for NER and RE fine-tuning ChatGLM2 and Alpaca2.}
    \label{instruction}

\end{table*}
\section{Experiments}
\subsection{Experimental Setting}
We model NER as a sequence labeling task and RE as a relation classification task.
We select baseline models from different paradigms to assess the challenges of information extraction in ancient Chinese historical documents. In recent studies, researchers~\cite{Wang_Zhou_Zu_Xia_Chen_Zhang_Zheng_Ye_Zhang_Gui_et_2023, Gui_Zhang_Ye_Zhang_2023} have achieved success using LLMs for information extraction. Therefore, we choose both open-source and closed-source LLMs as our baseline models. Our experiments involve techniques such as In-Context Learning~\cite{Brown_Mann_Ryder_Subbiah_Kaplan_Dhariwal_Neelakantan_Shyam_Sastry_Askell_et_2020}, LoRA~\cite{Hu_Shen_Wallis_Allen-Zhu_Li_Wang_Wang_Chen_2021}, P-tuning~\cite{Liu_Ji_Fu_Tam_Du_Yang_Tang_2022}, and Fine-Tuning on language models of various sizes.

Our baseline models are as follows.

\begin{itemize}
    \item SikuBERT~\footnote{\url{https://huggingface.co/SIKU-BERT/sikubert}}: a BERT model that has been incrementally trained with an ancient Chinese corpus. In our approach, we utilize SikuBERT as an encoder for both the NER task, where we used CRF as the decoder, and the RE task, for which we employed MLP+softmax as the classifier.
    
    \item SikuRoBERTa~\footnote{\url{https://huggingface.co/SIKU-BERT/sikuroberta}}:  a RoBERTa model that has been incrementally trained with an ancient Chinese corpus. All experiments are conducted in a manner similar to those using SikuBERT.
    
    \item ChatGLM2-6B~\footnote{\url{https://huggingface.co/THUDM/chatglm2-6b}}: an open-source bilingual (Chinese-English) chat model. We fine-tune it using the P-Tuning v2 technique~\cite{DBLP:journals/corr/abs-2110-07602}. Our training samples, as shown in Table~\ref{instruction}, include three components: task description, input, and output.

    \item Alpaca2-7B~\footnote{\url{https://github.com/ymcui/Chinese-LLaMA-Alpaca-2}}: 
    a model based on LLaMA2~\cite{Touvron_Martin_Stone_Albert_Almahairi_Babaei_Bashlykov_Batra_Bhargava_Bhosale_et_2023}, and it has been further pre-trained on a Chinese corpus. We fine-tuned it using LoRA~\cite{Hu_Shen_Wallis_Allen-Zhu_Li_Wang_Wang_Chen_2021}, following the same training samples as for ChatGLM2-6B.

    \item GPT3.5: a large language model with approximately 200B parameters. Due to its closed-source nature, fine-tuning was performed exclusively through the In-Context Learning method. For the NER task, we select 5 random examples from the training set as demonstrations. In the case of the RE task, we draw one sample from the training set for each relation type as demonstrations, i.e., 12-way 1-shot. 
    
\end{itemize}

\subsection{NER Experimental Results and Analysis}

The experimental results for named entity recognition are shown in Table~\ref{ner_result}. We report micro F1, macro F1, and the model's performance on each entity type.


From the NER experimental results, it's clear that PLMs outperform LLMs. This could be due to two possible factors. First, PLMs are incrementally trained in Ancient Chinese, giving them a superior understanding of this language. Whereas ChatGLM2 and Alpaca2 may have a small percentage of Ancient Chinese in the training corpus, therefore, the Ancient Chinese understanding ability of them is inferior to that of PLMs. Second, fine-tuning, which involves adjusting all model parameters, appears to be more effective than the partial modifications made by LoRA and P-tuning. Remarkably, GPT-3.5 shows promising results with just five examples, suggesting the potential of in-context learning in the NER task with this model.

In addition, we observe the performance of models on each entity class and find that the two classes of entities, Official and Book, are relatively ineffective, which is probably because, officials are era-specific, with different names for different dynasties, while books may be due to insufficient training data.

\begin{table*}[h]
\small
\centering
\setlength{\tabcolsep}{1mm}
\begin{tabular}{c|c|c|c|c|c|c|c} 
\Xhline{1.2pt}
\hline
\textbf{Techniques}                           &\textbf{ Models }                                 & \multicolumn{1}{c|}{\textbf{Entity Type}} & \multicolumn{1}{c|}{\textbf{P}}     & \multicolumn{1}{c|}{\textbf{R}}     & \multicolumn{1}{c|}{\textbf{F1} }    & \textbf{MacroF1 }               & \textbf{MicroF1}                 \\ 
\hline
\multirow{18}{*}{Finu-tuning}        & \multicolumn{7}{c}{Ancient Chinese fine-tuned Models}                                                                                                                                                              \\ 
\cline{2-8}
                                     & \multirow{4}{*}{SikuBERT}               & PER                              & 94.87                           &  96.74                          &  95.66                          & \multirow{4}{*}{88.67}      & \multirow{4}{*}{92.31}       \\ 
\cline{3-6}
                                     &                                         & LOC                              &  92.76                        & 93.41                           & 93.08                           &                        &                         \\ 
\cline{3-6}
                                     &                                         & OFI                              &   83.94                         & 84.97                           &    84.45                        &                        &                         \\ 
\cline{3-6}
                                     &                                         & BOOK                             &   78.57                         &    84.62                        &   81.48                         &                        &                         \\ 
\cline{2-8}
                                     & \multirow{4}{*}{SikuRoBERTa}            & PER                              &  94.87                          &     96.47                       &   95.66                         & \multirow{4}{*}{\textbf{90.47} }     & \multirow{4}{*}{\textbf{92.46}}       \\ 
\cline{3-6}
                                     &                                         & LOC                              &  92.47                          &  92.47                          & 92.47                           &                        &                         \\ 
\cline{3-6}
                                     &                                         & OFI                              & 83.87                           & 87.73                           &  85.76                          &                        &                         \\ 
\cline{3-6}
                                     &                                         & BOOK                             & 91.67                           &   84.62                         &   88.00                         &                        &                         \\ 
\cline{2-8}
                                     & \multicolumn{7}{c}{Large Language Models}                                                                                                                                                                   \\ 
\cline{2-8}
                                     & \multirow{4}{*}{ChatGLM2 (6B, P-tuning)} & PER                              & \multicolumn{1}{r|}{84.57} & \multicolumn{1}{r|}{78.62} & \multicolumn{1}{r|}{81.49} & \multirow{4}{*}{64.90} & \multirow{4}{*}{74.89}  \\ 
\cline{3-6}
                                     &                                         & LOC                              & \multicolumn{1}{r|}{77.11} & \multicolumn{1}{r|}{72.94} & \multicolumn{1}{r|}{74.97} &                        &                         \\ 
\cline{3-6}
                                     &                                         & OFI                              & \multicolumn{1}{r|}{64.75} & \multicolumn{1}{r|}{58.59} & \multicolumn{1}{r|}{61.51} &                        &                         \\ 
\cline{3-6}
                                     &                                         & BOOK                             & \multicolumn{1}{r|}{45.45} & \multicolumn{1}{r|}{38.46} & \multicolumn{1}{r|}{41.67} &                        &                         \\ 
\cline{2-8}
                                     & \multirow{4}{*}{Alpaca2 (7B, LoRA)}      & PER                              & \multicolumn{1}{r|}{90.29} & \multicolumn{1}{r|}{87.62} & \multicolumn{1}{r|}{88.94} & \multirow{4}{*}{78.15} & \multirow{4}{*}{85.83}  \\ 
\cline{3-6}
                                     &                                         & LOC                              & \multicolumn{1}{r|}{89.10} & \multicolumn{1}{r|}{86.59} & \multicolumn{1}{r|}{87.83} &                        &                         \\ 
\cline{3-6}
                                     &                                         & OFI                              & \multicolumn{1}{r|}{78.37} & \multicolumn{1}{r|}{76.69}  & \multicolumn{1}{r|}{77.52} &                        &                         \\ 
\cline{3-6}
                                     &                                         & BOOK                             & \multicolumn{1}{r|}{63.64}    & \multicolumn{1}{r|}{53.85} & \multicolumn{1}{r|}{58.33} &                        &                         \\ 
\hline
\multirow{4}{*}{In-Context Learning} & \multirow{4}{*}{GPT-3.5 (5 shot)}       & PER                              &  55.26                          &   61.47                         & 58.20                           & \multirow{4}{*}{45.34}      & \multirow{4}{*}{56.72}       \\ 
\cline{3-6}
                                     &                                         & LOC                              &  52.73                          &      60.83                      &      56.49                     &                        &                         \\ 
\cline{3-6}
                                     &                                         & OFI                              &       54.29                     & 59.79                           & 56.91                           &                        &                         \\ 
\cline{3-6}
                                     &                                         & BOOK                             &  9.52                          &          10.00                  & 9.76                           &                        &                         \\
\hline
\Xhline{1.2pt}
\end{tabular}
\caption{Experimental results for the NER task. Both the macro F1 (\%) and the micro F1 (\%) are evaluation metrics. We highlight the highest performance in bold.}
\label{ner_result}
\end{table*}

\begin{table*}[]
   \small
   \setlength{\tabcolsep}{4mm}
    \centering
    \begin{tabular}{c|c|c|c|c|c}
    \hline
    \Xhline{1.2pt}
        \multirow{3}{*}{\textbf{Techniques}}&\multirow{3}{*}{\textbf{Models}} & \multicolumn{4}{c}{\textbf{RE}}\\
        \cline{3-6}
       & & \multicolumn{3}{c|}{\textbf{Macro}} &\multirow{2}{*}{\textbf{Acc.}}  \\
        \cline{3-5}
         && \textbf{P} & \textbf{R} &  \textbf{F1} &  \\
        \cline{1-6}
       \multirow{6}{*}{Fine-tuning}& \multicolumn{5}{c}{Ancient Chinese fine-tuned Models}\\
         \cline{2-6}
       
        & SikuBERT &83.89 & 82.48& 83.18 & 88.47\\
         \cline{2-6}
         &SikuRoBERTa &79.93&79.98& 79.95&87.45\\
         \cline{2-6}
         & \multicolumn{5}{c}{Large Language Models}\\
         \cline{2-6}
        &ChatGLM2 (6B, P-tuning) &76.14&76.56&76.35&83.86\\
         \cline{2-6}
         
        & Alpaca2 (7B, LoRA) &\textbf{84.53}&\textbf{85.46}&\textbf{85.00}&\textbf{89.48}\\
        \cline{1-6}
   \multirow{1}{*}{In-Context Learning}  & GPT-3.5 (12-way 1-shot)  & 40.04 & 14.27 &21.04 & 53.74\\
        \cline{1-6}

         \Xhline{1.2pt}
    \end{tabular}
    \caption{Experimental results for the RE task. Both the macro F1 (\%) and accuracy (Acc. \%) are evaluation metrics. We highlight the best performance in bold.}
    \label{re_result}
\end{table*}

\subsection{RE Experimental Results and Analysis}
Table~\ref{re_result} lists the experimental results for the RE task, where we also report micro F1 and accuracy.

By analyzing the results, we find that the performance of ChatGLM2 and Alpaca2 is comparable to that of PLMs. Even Alpaca2 achieved the best performance. As for GPT-3.5, the limited number of samples provided seems to hinder its ability to confine relation types to the predefined set, resulting in the generation of relation types not within the set and, consequently, lower recall rates. It's possible that clearer, more comprehensive, and richer prompts could yield improved experimental results.

Due to space constraints, we don't report the F1 scores for each relation type, but we plan to add them in the appendix in the future. It's worth noting that all models exhibit relatively poor performance in one specific relationship type: Political support. The experimental results for each model in the ``Political support'' relation are presented in Table~\ref{each_type}. We speculate that this could be attributed to the semantic complexity of this relation type, which includes terms such as ``support'', ``recommend'', and ``rescue''. That makes it challenging for the models to summarize the features associated with this specific relation.

\begin{table}[]
    \centering
    \small
    \begin{tabular}{c|c|c|c}
        \Xhline{1.2pt}
        \hline
         \multirow{2}{*}{\textbf{Models}}  &\multicolumn{3}{c}{\textbf{Political support}} \\
        \cline{2-4}
         & \textbf{P}  &  \textbf{R}  &  \textbf{F} \\
         \hline
         
        SikuBERT &61.82& 61.82 & 61.82 \\
         \hline
        SikuRoBERTa &52.83 &50.91 & 51.85\\
         \hline
       ChatGLM2 (6B) & 56.25	&49.09	&52.43 \\
        \hline
       Alpaca2 (7B) & 70.59 & 65.45 &67.92  \\
        \hline
       GPT-3.5 (12-way 1-shot) &14.29 &1.82  &3.23 \\
        \hline
        \Xhline{1.2pt}
    \end{tabular}
    \caption{The performance of all models on the Political support relation type.}
    \label{each_type}
\end{table}

In summary, pre-trained language models remain highly potent base models when ample training data is available. It's important to note that large language models exhibit substantial performance variations across different tasks. For instance, ChatGLM2 and Alpaca2 demonstrate superior performance in the RE task as opposed to the NER task. This can be attributed to the NER task's greater demand for polyglot features from the model, including entity position identification and entity type recognition. In contrast, the RE task shares similarities with sentence classification, making it a more manageable challenge for these large language models.

\section{Conclusion}
In this paper, we propose CHisIEC, an ancient Chinese history corpus for NER and RE tasks. Our dataset contains texts from 13 dynasties, epitomizing the extensive temporal scope and text heterogeneity of Chinese historical literature. We conduct experiments on both the pre-trained language models and the large language models to validate the applicability of the dataset, and also evaluate the capability of the LLMs in the domain tasks of ancient Chinese history.
%
\section{Acknowledgments}
This research is supported by the NSFC project “the Construction of the Knowledge Graph for the History of Chinese Confucianism” (Grant No. 72010107003). 

\section{Bibliographical References}

\bibliographystyle{lrec-coling2024-natbib}
\bibliography{lrec-coling2024-example}

\begin{thebibliography}{31}
\expandafter\ifx\csname natexlab\endcsname\relax\def\natexlab#1{#1}\fi

\bibitem[{Bekele et~al.(2016)Bekele, De~By, and
  Singh}]{Bekele_De_By_Singh_2016}
Mafkereseb Bekele, Rolf De~By, and Gaurav Singh. 2016.
\newblock \href {https://doi.org/10.3390/ijgi5120221} {Spatiotemporal
  information extraction from a historic expedition gazetteer}.
\newblock \emph{ISPRS International Journal of Geo-Information}, 5(12):221.

\bibitem[{Brown et~al.(2020{\natexlab{a}})Brown, Mann, Ryder, Subbiah, Kaplan,
  Dhariwal, Neelakantan, Shyam, Sastry, Askell, Agarwal, Herbert-Voss, Krueger,
  Henighan, Child, Ramesh, Ziegler, Wu, Winter, Hesse, Chen, Sigler, Litwin,
  Gray, Chess, Clark, Berner, McCandlish, Radford, Sutskever, and
  Amodei}]{Brown_2020}
Tom~B. Brown, Benjamin Mann, Nick Ryder, Melanie Subbiah, Jared Kaplan,
  Prafulla Dhariwal, Arvind Neelakantan, Pranav Shyam, Girish Sastry, Amanda
  Askell, Sandhini Agarwal, Ariel Herbert-Voss, Gretchen Krueger, Tom Henighan,
  Rewon Child, Aditya Ramesh, Daniel~M. Ziegler, Jeffrey Wu, Clemens Winter,
  Christopher Hesse, Mark Chen, Eric Sigler, Mateusz Litwin, Scott Gray,
  Benjamin Chess, Jack Clark, Christopher Berner, Sam McCandlish, Alec Radford,
  Ilya Sutskever, and Dario Amodei. 2020{\natexlab{a}}.
\newblock \href {http://arxiv.org/abs/2005.14165} {Language models are few-shot
  learners}.
\newblock (arXiv:2005.14165).
\newblock ArXiv:2005.14165 [cs].

\bibitem[{Brown et~al.(2020{\natexlab{b}})Brown, Mann, Ryder, Subbiah, Kaplan,
  Dhariwal, Neelakantan, Shyam, Sastry, Askell, Agarwal, Herbert-Voss, Krueger,
  Henighan, Child, Ramesh, Ziegler, Wu, Winter, Hesse, Chen, Sigler, Litwin,
  Gray, Chess, Clark, Berner, McCandlish, Radford, Sutskever, and
  Amodei}]{Brown_Mann_Ryder_Subbiah_Kaplan_Dhariwal_Neelakantan_Shyam_Sastry_Askell_et_2020}
Tom~B. Brown, Benjamin Mann, Nick Ryder, Melanie Subbiah, Jared Kaplan,
  Prafulla Dhariwal, Arvind Neelakantan, Pranav Shyam, Girish Sastry, Amanda
  Askell, Sandhini Agarwal, Ariel Herbert-Voss, Gretchen Krueger, Tom Henighan,
  Rewon Child, Aditya Ramesh, Daniel~M. Ziegler, Jeffrey Wu, Clemens Winter,
  Christopher Hesse, Mark Chen, Eric Sigler, Mateusz Litwin, Scott Gray,
  Benjamin Chess, Jack Clark, Christopher Berner, Sam McCandlish, Alec Radford,
  Ilya Sutskever, and Dario Amodei. 2020{\natexlab{b}}.
\newblock \href {http://arxiv.org/abs/2005.14165} {Language models are few-shot
  learners}.
\newblock (arXiv:2005.14165).
\newblock ArXiv:2005.14165 [cs].

\bibitem[{Eberts and Ulges(2021)}]{Eberts_Ulges_2021}
Markus Eberts and Adrian Ulges. 2021.
\newblock \href {https://doi.org/10.3233/FAIA200321} {Span-based joint entity
  and relation extraction with transformer pre-training}.
\newblock ArXiv:1909.07755 [cs].

\bibitem[{Ehrmann et~al.(2020)Ehrmann, Romanello, Fluckiger, and
  Clematide}]{Ehrmann_Romanello_Fluckiger_Clematide}
Maud Ehrmann, Matteo Romanello, Alex Fluckiger, and Simon Clematide. 2020.
\newblock Extended overview of clef hipe 2020: Named entity processing on
  historical newspapers.

\bibitem[{Gui et~al.(2023)Gui, Zhang, Ye, and Zhang}]{Gui_Zhang_Ye_Zhang_2023}
Honghao Gui, Jintian Zhang, Hongbin Ye, and Ningyu Zhang. 2023.
\newblock \href {http://arxiv.org/abs/2305.11527} {Instructie: A chinese
  instruction-based information extraction dataset}.
\newblock (arXiv:2305.11527).
\newblock ArXiv:2305.11527 [cs].

\bibitem[{Hu et~al.(2021)Hu, Shen, Wallis, Allen-Zhu, Li, Wang, Wang, and
  Chen}]{Hu_Shen_Wallis_Allen-Zhu_Li_Wang_Wang_Chen_2021}
Edward~J. Hu, Yelong Shen, Phillip Wallis, Zeyuan Allen-Zhu, Yuanzhi Li, Shean
  Wang, Lu~Wang, and Weizhu Chen. 2021.
\newblock \href {http://arxiv.org/abs/2106.09685} {Lora: Low-rank adaptation of
  large language models}.
\newblock (arXiv:2106.09685).
\newblock ArXiv:2106.09685 [cs].

\bibitem[{Huguet~Cabot and Navigli(2021)}]{Huguet_Cabot_Navigli_2021}
Pere-Lluís Huguet~Cabot and Roberto Navigli. 2021.
\newblock \href {https://doi.org/10.18653/v1/2021.findings-emnlp.204} {Rebel:
  Relation extraction by end-to-end language generation}.
\newblock In \emph{Findings of the Association for Computational Linguistics:
  EMNLP 2021}, page 2370–2381, Punta Cana, Dominican Republic. Association
  for Computational Linguistics.

\bibitem[{Ji et~al.(2020)Ji, Yu, Li, Ma, Wu, Tan, and
  Liu}]{Ji_Yu_Li_Ma_Wu_Tan_Liu_2020}
Bin Ji, Jie Yu, Shasha Li, Jun Ma, Qingbo Wu, Yusong Tan, and Huijun Liu. 2020.
\newblock \href {https://doi.org/10.18653/v1/2020.coling-main.8} {Span-based
  joint entity and relation extraction with attention-based span-specific and
  contextual semantic representations}.
\newblock In \emph{Proceedings of the 28th International Conference on
  Computational Linguistics}, page 88–99, Barcelona, Spain (Online).
  International Committee on Computational Linguistics.

\bibitem[{Ji et~al.(2021)Ji, Chen, Han, and Wang}]{jizijing2021}
Zijing Ji, Zirui Chen, Lifan Han, and Xin Wang. 2021.
\newblock Research on information extraction methods for historical classics
  under the perspective of digital humanities.
\newblock \emph{Big Data Research}, pages 1--21.

\bibitem[{Kim et~al.(2022)Kim, Kim, Son, and Lim}]{Kim_Kim_Son_Lim_2022}
Gyeongmin Kim, Jinsung Kim, Junyoung Son, and Heuiseok Lim. 2022.
\newblock \href {https://aclanthology.org/2022.coling-1.308} {Kochet: A korean
  cultural heritage corpus for entity-related tasks}.
\newblock In \emph{Proceedings of the 29th International Conference on
  Computational Linguistics}, page 3496–3505, Gyeongju, Republic of Korea.
  International Committee on Computational Linguistics.

\bibitem[{Li et~al.(2021)Li, Wei, Liu, Chen, Fang, and
  Jiang}]{Li_Wei_Liu_Chen_Fang_Jiang_2021}
Bo~Li, Jiyu Wei, Yang Liu, Yuze Chen, Xi~Fang, and Bin Jiang. 2021.
\newblock \href {https://doi.org/10.3390/app112412060} {Few-shot relation
  extraction on ancient chinese documents}.
\newblock \emph{Applied Sciences}, 11(24):12060.

\bibitem[{Li et~al.(2019)Li, Yin, Sun, Li, Yuan, Chai, Zhou, and
  Li}]{Li_Yin_Sun_Li_Yuan_Chai_Zhou_Li_2019}
Xiaoya Li, Fan Yin, Zijun Sun, Xiayu Li, Arianna Yuan, Duo Chai, Mingxin Zhou,
  and Jiwei Li. 2019.
\newblock \href {https://doi.org/10.18653/v1/P19-1129} {Entity-relation
  extraction as multi-turn question answering}.
\newblock In \emph{Proceedings of the 57th Annual Meeting of the Association
  for Computational Linguistics}, page 1340–1350, Florence, Italy.
  Association for Computational Linguistics.

\bibitem[{Liang et~al.(2022)Liang, Wu, Li, and Li}]{Liang_Wu_Li_Li_2022}
Xinnian Liang, Shuangzhi Wu, Mu~Li, and Zhoujun Li. 2022.
\newblock \href {https://doi.org/10.18653/v1/2022.naacl-main.375} {Modeling
  multi-granularity hierarchical features for relation extraction}.
\newblock In \emph{Proceedings of the 2022 Conference of the North American
  Chapter of the Association for Computational Linguistics: Human Language
  Technologies}, page 5088–5098, Seattle, United States. Association for
  Computational Linguistics.

\bibitem[{Ling et~al.(2023)Ling, Zhao, Zhang, Liu, Cheng, Wang, Chen, Osaki,
  Matsuda, Chen, and
  Zhao}]{Ling_Zhao_Zhang_Liu_Cheng_Wang_Chen_Osaki_Matsuda_Chen_2023}
Chen Ling, Xujiang Zhao, Xuchao Zhang, Yanchi Liu, Wei Cheng, Haoyu Wang,
  Zhengzhang Chen, Takao Osaki, Katsushi Matsuda, Haifeng Chen, and Liang Zhao.
  2023.
\newblock \href {http://arxiv.org/abs/2309.03433} {Improving open information
  extraction with large language models: A study on demonstration uncertainty}.
\newblock (arXiv:2309.03433).
\newblock ArXiv:2309.03433 [cs].

\bibitem[{Liu et~al.(2021)Liu, Ji, Fu, Du, Yang, and
  Tang}]{DBLP:journals/corr/abs-2110-07602}
Xiao Liu, Kaixuan Ji, Yicheng Fu, Zhengxiao Du, Zhilin Yang, and Jie Tang.
  2021.
\newblock \href {http://arxiv.org/abs/2110.07602} {P-tuning v2: Prompt tuning
  can be comparable to fine-tuning universally across scales and tasks}.
\newblock \emph{CoRR}, abs/2110.07602.

\bibitem[{Liu et~al.(2022)Liu, Ji, Fu, Tam, Du, Yang, and
  Tang}]{Liu_Ji_Fu_Tam_Du_Yang_Tang_2022}
Xiao Liu, Kaixuan Ji, Yicheng Fu, Weng Tam, Zhengxiao Du, Zhilin Yang, and Jie
  Tang. 2022.
\newblock \href {https://doi.org/10.18653/v1/2022.acl-short.8} {P-tuning:
  Prompt tuning can be comparable to fine-tuning across scales and tasks}.
\newblock In \emph{Proceedings of the 60th Annual Meeting of the Association
  for Computational Linguistics (Volume 2: Short Papers)}, page 61–68,
  Dublin, Ireland. Association for Computational Linguistics.

\bibitem[{Lou et~al.(2023)Lou, Zhang, and Yin}]{Lou_Zhang_Yin_2023}
Renze Lou, Kai Zhang, and Wenpeng Yin. 2023.
\newblock \href {http://arxiv.org/abs/2303.10475} {Is prompt all you need? no.
  a comprehensive and broader view of instruction learning}.
\newblock (arXiv:2303.10475).
\newblock ArXiv:2303.10475 [cs].

\bibitem[{Nayak and Ng(2020)}]{Nayak_Ng_2020}
Tapas Nayak and Hwee~Tou Ng. 2020.
\newblock \href {https://doi.org/10.1609/aaai.v34i05.6374} {Effective modeling
  of encoder-decoder architecture for joint entity and relation extraction}.
\newblock \emph{Proceedings of the AAAI Conference on Artificial Intelligence},
  34(05):8528–8535.

\bibitem[{Neudecker(2016)}]{neudecker-2016-open}
Clemens Neudecker. 2016.
\newblock \href {https://aclanthology.org/L16-1689} {An open corpus for named
  entity recognition in historic newspapers}.
\newblock In \emph{Proceedings of the Tenth International Conference on
  Language Resources and Evaluation ({LREC}'16)}, pages 4348--4352,
  Portoro{\v{z}}, Slovenia. European Language Resources Association (ELRA).

\bibitem[{Nguyen et~al.(2022)Nguyen, Min, Dernoncourt, and
  Nguyen}]{Nguyen_Min_Dernoncourt_Nguyen_2022}
Minh~Van Nguyen, Bonan Min, Franck Dernoncourt, and Thien Nguyen. 2022.
\newblock \href {https://doi.org/10.18653/v1/2022.naacl-main.324} {Joint
  extraction of entities, relations, and events via modeling inter-instance and
  inter-label dependencies}.
\newblock In \emph{Proceedings of the 2022 Conference of the North American
  Chapter of the Association for Computational Linguistics: Human Language
  Technologies}, page 4363–4374, Seattle, United States. Association for
  Computational Linguistics.

\bibitem[{Nundloll et~al.(2022)Nundloll, Smail, Stevens, and
  Blair}]{Nundloll_Smail_Stevens_Blair_2022}
Vatsala Nundloll, Robert Smail, Carly Stevens, and Gordon Blair. 2022.
\newblock \href {https://doi.org/https://doi.org/10.1016/j.heliyon.2022.e10710}
  {Automating the extraction of information from a historical text and building
  a linked data model for the domain of ecology and conservation science}.
\newblock \emph{Heliyon}, 8(10):e10710.

\bibitem[{Ouyang et~al.(2022)Ouyang, Wu, Jiang, Almeida, Wainwright, Mishkin,
  Zhang, Agarwal, Slama, Ray, Schulman, Hilton, Kelton, Miller, Simens, Askell,
  Welinder, Christiano, Leike, and
  Lowe}]{Ouyang_Wu_Jiang_Almeida_Wainwright_Mishkin_Zhang_Agarwal_Slama_Ray_et_2022}
Long Ouyang, Jeff Wu, Xu~Jiang, Diogo Almeida, Carroll~L. Wainwright, Pamela
  Mishkin, Chong Zhang, Sandhini Agarwal, Katarina Slama, Alex Ray, John
  Schulman, Jacob Hilton, Fraser Kelton, Luke Miller, Maddie Simens, Amanda
  Askell, Peter Welinder, Paul Christiano, Jan Leike, and Ryan Lowe. 2022.
\newblock \href {http://arxiv.org/abs/2203.02155} {Training language models to
  follow instructions with human feedback}.
\newblock (arXiv:2203.02155).
\newblock ArXiv:2203.02155 [cs].

\bibitem[{Touvron et~al.(2023)Touvron, Martin, Stone, Albert, Almahairi,
  Babaei, Bashlykov, Batra, Bhargava, Bhosale, Bikel, Blecher, Ferrer, Chen,
  Cucurull, Esiobu, Fernandes, Fu, Fu, Fuller, Gao, Goswami, Goyal, Hartshorn,
  Hosseini, Hou, Inan, Kardas, Kerkez, Khabsa, Kloumann, Korenev, Koura,
  Lachaux, Lavril, Lee, Liskovich, Lu, Mao, Martinet, Mihaylov, Mishra,
  Molybog, Nie, Poulton, Reizenstein, Rungta, Saladi, Schelten, Silva, Smith,
  Subramanian, Tan, Tang, Taylor, Williams, Kuan, Xu, Yan, Zarov, Zhang, Fan,
  Kambadur, Narang, Rodriguez, Stojnic, Edunov, and
  Scialom}]{Touvron_Martin_Stone_Albert_Almahairi_Babaei_Bashlykov_Batra_Bhargava_Bhosale_et_2023}
Hugo Touvron, Louis Martin, Kevin Stone, Peter Albert, Amjad Almahairi, Yasmine
  Babaei, Nikolay Bashlykov, Soumya Batra, Prajjwal Bhargava, Shruti Bhosale,
  Dan Bikel, Lukas Blecher, Cristian~Canton Ferrer, Moya Chen, Guillem
  Cucurull, David Esiobu, Jude Fernandes, Jeremy Fu, Wenyin Fu, Brian Fuller,
  Cynthia Gao, Vedanuj Goswami, Naman Goyal, Anthony Hartshorn, Saghar
  Hosseini, Rui Hou, Hakan Inan, Marcin Kardas, Viktor Kerkez, Madian Khabsa,
  Isabel Kloumann, Artem Korenev, Punit~Singh Koura, Marie-Anne Lachaux,
  Thibaut Lavril, Jenya Lee, Diana Liskovich, Yinghai Lu, Yuning Mao, Xavier
  Martinet, Todor Mihaylov, Pushkar Mishra, Igor Molybog, Yixin Nie, Andrew
  Poulton, Jeremy Reizenstein, Rashi Rungta, Kalyan Saladi, Alan Schelten, Ruan
  Silva, Eric~Michael Smith, Ranjan Subramanian, Xiaoqing~Ellen Tan, Binh Tang,
  Ross Taylor, Adina Williams, Jian~Xiang Kuan, Puxin Xu, Zheng Yan, Iliyan
  Zarov, Yuchen Zhang, Angela Fan, Melanie Kambadur, Sharan Narang, Aurelien
  Rodriguez, Robert Stojnic, Sergey Edunov, and Thomas Scialom. 2023.
\newblock \href {http://arxiv.org/abs/2307.09288} {Llama 2: Open foundation and
  fine-tuned chat models}.
\newblock (arXiv:2307.09288).
\newblock ArXiv:2307.09288 [cs].

\bibitem[{Wang et~al.(2023)Wang, Zhou, Zu, Xia, Chen, Zhang, Zheng, Ye, Zhang,
  Gui, Kang, Yang, Li, and
  Du}]{Wang_Zhou_Zu_Xia_Chen_Zhang_Zheng_Ye_Zhang_Gui_et_2023}
Xiao Wang, Weikang Zhou, Can Zu, Han Xia, Tianze Chen, Yuansen Zhang, Rui
  Zheng, Junjie Ye, Qi~Zhang, Tao Gui, Jihua Kang, Jingsheng Yang, Siyuan Li,
  and Chunsai Du. 2023.
\newblock \href {http://arxiv.org/abs/2304.08085} {Instructuie: Multi-task
  instruction tuning for unified information extraction}.
\newblock (arXiv:2304.08085).
\newblock ArXiv:2304.08085 [cs].

\bibitem[{Wei et~al.(2023)Wei, Cui, Cheng, Wang, Zhang, Huang, Xie, Xu, Chen,
  Zhang, Jiang, and
  Han}]{Wei_Cui_Cheng_Wang_Zhang_Huang_Xie_Xu_Chen_Zhang_et_2023}
Xiang Wei, Xingyu Cui, Ning Cheng, Xiaobin Wang, Xin Zhang, Shen Huang, Pengjun
  Xie, Jinan Xu, Yufeng Chen, Meishan Zhang, Yong Jiang, and Wenjuan Han. 2023.
\newblock \href {http://arxiv.org/abs/2302.10205} {Zero-shot information
  extraction via chatting with chatgpt}.
\newblock (arXiv:2302.10205).
\newblock ArXiv:2302.10205 [cs].

\bibitem[{Wu and He(2019)}]{Wu_He_2019}
Shanchan Wu and Yifan He. 2019.
\newblock \href {http://arxiv.org/abs/1905.08284} {Enriching pre-trained
  language model with entity information for relation classification}.
\newblock (arXiv:1905.08284).
\newblock ArXiv:1905.08284 [cs].

\bibitem[{Zhao et~al.(2020)Zhao, Yan, Cao, and Li}]{Zhao_Yan_Cao_Li_2020}
Tianyang Zhao, Zhao Yan, Yunbo Cao, and Zhoujun Li. 2020.
\newblock \href {https://doi.org/10.24963/ijcai.2020/546} {Asking effective and
  diverse questions: A machine reading comprehension based framework for joint
  entity-relation extraction}.
\newblock In \emph{Proceedings of the Twenty-Ninth International Joint
  Conference on Artificial Intelligence}, page 3948–3954, Yokohama, Japan.
  International Joint Conferences on Artificial Intelligence Organization.

\bibitem[{Zhao et~al.(2023)Zhao, Zhou, Li, Tang, Wang, Hou, Min, Zhang, Zhang,
  Dong, Du, Yang, Chen, Chen, Jiang, Ren, Li, Tang, Liu, Liu, Nie, and
  Wen}]{Zhao_Zhou_Li_Tang_Wang_Hou_Min_Zhang_Zhang_Dong_et_2023}
Wayne~Xin Zhao, Kun Zhou, Junyi Li, Tianyi Tang, Xiaolei Wang, Yupeng Hou,
  Yingqian Min, Beichen Zhang, Junjie Zhang, Zican Dong, Yifan Du, Chen Yang,
  Yushuo Chen, Zhipeng Chen, Jinhao Jiang, Ruiyang Ren, Yifan Li, Xinyu Tang,
  Zikang Liu, Peiyu Liu, Jian-Yun Nie, and Ji-Rong Wen. 2023.
\newblock \href {http://arxiv.org/abs/2303.18223} {A survey of large language
  models}.
\newblock (arXiv:2303.18223).
\newblock ArXiv:2303.18223 [cs].

\bibitem[{Zheng et~al.(2017)Zheng, Wang, Bao, Hao, Zhou, and
  Xu}]{Zheng_Wang_Bao_Hao_Zhou_Xu_2017}
Suncong Zheng, Feng Wang, Hongyun Bao, Yuexing Hao, Peng Zhou, and Bo~Xu. 2017.
\newblock \href {https://doi.org/10.18653/v1/P17-1113} {Joint extraction of
  entities and relations based on a novel tagging scheme}.
\newblock In \emph{Proceedings of the 55th Annual Meeting of the Association
  for Computational Linguistics (Volume 1: Long Papers)}, page 1227–1236,
  Vancouver, Canada. Association for Computational Linguistics.

\bibitem[{Zinin and Xu(2020)}]{Zinin_Xu}
Sergey Zinin and Yang Xu. 2020.
\newblock Corpus of chinese dynastic histories: Gender analysis over two
  millennia.
\newblock pages 785--793.

\end{thebibliography}

\end{CJK*}

\end{document}